\documentclass[runningheads]{llncs}
\usepackage[T1]{fontenc}
\usepackage{graphicx}
\usepackage{amsmath, amssymb}
\usepackage{makecell}
\usepackage{multirow}
\usepackage{booktabs}
\usepackage{comment}
\usepackage{hyperref}
\usepackage{algorithm}
\usepackage{algorithmic}
\newlength\myindent
\begin{document}

\title{SMC-UDA: Structure-Modal Constraint for Unsupervised Cross-Domain Renal Segmentation}
\titlerunning{SMC-UDA: Cross-Domain Renal Segmentation}

% \author{Anonymized MICCAI submission}
% \institute{~}
\author{Zhusi Zhong\inst{1,2,3}, Jie Li\inst{1}, Lulu Bi\inst{2,3}, Li Yang\inst{4}, Ihab Kamel\inst{5}, Rama Chellappa\inst{5}, Xinbo Gao\inst{1}, Harrison Bai\inst{5} and Zhicheng Jiao\inst{2,3} }

\authorrunning{F. Author et al.}
\institute{School of Electronic Engineering, Xidian University, Xi’an 710071, China \and 
Department of Diagnostic Radiology, Rhode Island Hospital, Providence 02903, USA \and
Warren Alpert Medical School of Brown University, Providence, 02903, USA
\email{zhicheng\_jiao@brown.edu} \and
School of Computer Science and Engineering, Central South University, Changsha 410083, China \and 
Department of Radiology and Radiological Sciences, Johns Hopkins University School of Medicine, Baltimore 21205, USA\\
}
\maketitle              % typeset the header of the contribution
\begin{abstract}
Medical image segmentation based on deep learning often fails when deployed on images from a different domain. The domain adaptation methods aim to solve domain-shift challenges, but still face some problems. The transfer learning methods require annotation on the target domain, and the generative unsupervised domain adaptation (UDA) models ignore domain-specific representations, whose generated quality highly restricts segmentation performance. In this study, we propose a novel \textbf{S}tructure-\textbf{M}odal \textbf{C}onstrained (SMC) UDA framework based on a discriminative paradigm and introduce edge structure as a bridge between domains. The proposed multi-modal learning backbone distills structure information from image texture to distinguish domain-invariant edge structure. With the structure-constrained self-learning and progressive ROI, our methods segment the kidney by locating the 3D spatial structure of the edge. We evaluated SMC-UDA on public renal segmentation datasets, adapting from the labeled source domain (CT) to the unlabeled target domain (CT/MRI). The experiments show that our proposed SMC-UDA has a strong generalization and outperforms generative UDA methods. 
\end{abstract}
\section{Introduction}
Image segmentation is a key component in medical image analysis, which supports disease diagnosis and quantification. Although fully supervised models with well-annotated data achieve promising performance, manual annotation is a highly tedious and time-consuming process, and generalization is impaired by the severe domain shifts due to different acquisitions between cases, institutions and modalities. Such challenges demonstrate the need for domain adaptation models, which address the problems posed by unlabeled data and domain shifts. Given two datasets from different domains, source domain $\mathit{D_{s}} = \{x_{s}^{img}, y_{s}^{seg}\}$ and target domain $\mathit{D_{t}} = \{x_{t}^{img}\}$ where $x^{img}$ and $y^{seg}$ represent a 3D image and an annotated mask, respectively. An intuitive transfer reuses and fine-tunes the pre-trained models for the new domains, but a large amount of target domain data and annotation $y_{t}^{seg}$ are necessary for the inductive training. In this regard, UDA methods without target domain annotations can effectively reduce data requirements.\par
The aim of current UDA methods is to learn a domain-invariant mapping relationship between $x_{s}^{img}$ and $x_{t}^{img}$. These methods employ pixel or feature alignment with adversarial training \cite{PnPAdaNet,sifa1,sifa2} to generate images in the other domain’s style. They train the segmentation module with the generated source images or indirectly segment the transferred target domain images. However, there exist limitations to this type of generative approaches. The most critical challenge is the unreliability of the synthetically generated images, making the model less convincing. Additionally, the generative UDA models learn the partial intensity projection while ignoring domain-specific representations, and they restrict UDA segmentation performance by the quality of the generated image. Moreover, the generative UDA models also require a substantial amount of training data for accurate mapping learning and generalization. Despite the aforementioned shortcomings due to computation limitations, most UDA methods process the 3D medical images using 2D techniques, which under-utilize spatial structure information. \par
Structure information in medical imaging refers to the spatial arrangements of tissues, organs, and other structures within the human body, and it is crucial for accurate disease detection, medical diagnosis, and surgical planning \cite{medsurgical,medRegistration}. It is an important aspect of the image information indicating tissue texture as well as spatial relationships between different structures, of various shapes and sizes. Although the image texture distributions vary between different domains, such as CT and MRI, the implicit structural information between modalities has high similarity and strong cross-domain invariance, as it is determined only by the physical spatial distribution. In cross-modal UDA \cite{jaritz2022cross}, the segmentation based on spatial structural information, instead of image texture translation, overcomes the challenge of shape detection on target domain through the effective constraint of the prior structural knowledge from the source domain. Additionally, processing integral 3D images instead of 2D slices provides a more comprehensive organ assessment, since 3D images preserve complete context information \cite{fard2022cnns}. The recent advances in point cloud methods \cite{qi2017pointnet++,tang2020searching,yan20222dpass,jaritz2022cross} have shown a strong ability to generalize 3D structure representations, making multi-modal learning for point cloud segmentation promising for cross-domain adaptation.\par
In this work, we propose a novel medical image segmentation framework, SMC-UDA (Structure-Modal Constrained Unsupervised Domain Adaption), and extend the multi-modal backbone to structure-constrained UDA. Our contributions lie in the following three points. (1) To the best of our knowledge, this is the first attempt to introduce the edge structure as a bridge between domains, and we propose a new segmentation paradigm that transforms the voxel segmentation to point cloud segmentation. (2) The backbone uses cross-modal distillation to enhance the structure representation and learns domain-invariant structural information from domain-specific image texture appearance. (3) A novel UDA training strategy with progressive ROI is developed to address the limited number of point clouds. The verified results on various public datasets on kidney segmentation demonstrate the effectiveness of our SMC-UDA.

\begin{figure}[!t]
\label{SMC-UDA_framework}
\centering
\includegraphics[scale=.41]{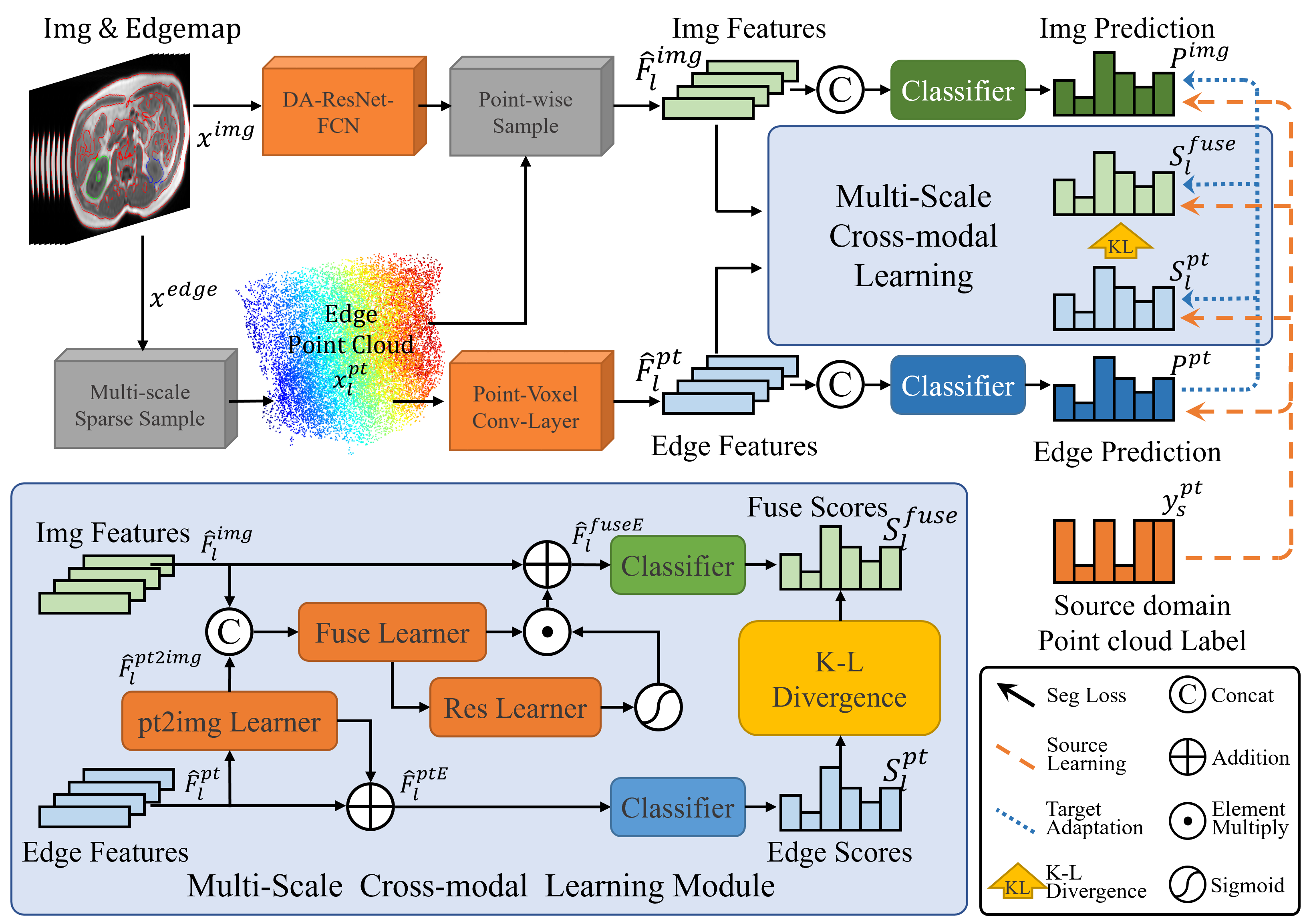}
\caption{Overview of the SMC-UDA framework, a multi-modal learning approach for medical image UDA segmentation. The multi-modal backbone includes two branches encoding image and edge features for point cloud segmentation and multi-modal learning (left). Supervised source domain training and self-supervised target domain training are indicated by the orange and blue dotted lines (right). For each scale of the Cross-Modal Knowledge Distillation learning (lower), the predictions are generated by the unshared MLP layers.}
\end{figure}

\section{Method}
\subsection{Multi-Modal Learning}
% \subsubsection{Edge Sparse Sampling:}
\textbf{Multi-Scale Feature Encoder.}
% Obtain edge by DoG. Sample point cloud on edge map. Do multi-modal point cloud segmentation. 
We obtain the edge map $x^{edge}$ for each image by applying the Marr-Hildreth operator \cite{edgedetectheory} in $x_{s}^{img}$ and $x_{t}^{img}$. The edge-based segmentation model requires a sufficient edge structure to perform point cloud segmentation. Although there is a direct mapping relationship between voxels and point clouds in 3D image, the edge voxels are too dense to compute with a large number of point clouds. We address this problem by randomly sampling $N$ positive edge voxels in $x^{edge}$ and $y^{seg}$ as the edge point cloud $x^{pt}\in R^{(N\times 3)}$ and its corresponding point cloud labels $y^{pt}\in [\![1,C]\!]^{N}$. \par
As shown in Fig. \ref{SMC-UDA_framework}, we utilize two networks to encode multi-scale features from 3D images and edge point clouds, respectively \cite{yan20222dpass}. The 3D ResNet34 serves as the image encoder to extract features $F_{l}^{img}$ and shares the parameters of deep Res-Layers in cross-domain. Meanwhile, a hierarchical encoder structure with sparse convolutional layers \cite{graham20183d} which constructed with point-voxel convolution and ResNet bottleneck \cite{tang2020searching}, is implied for extracting the point cloud features $F_l^{pt}$. The sparse convolution operation considers only non-empty voxels with high computational efficiency. Specifically, the two encoders extract $L$ feature maps from various scales, and the $l$-th features $F_{l}^{img} \in R^{H_{l}\times W_{l}\times D_{l}\times d_{l}}$ and $F_l^{pt}\in R^{N\times d_l}$ are upscaled back separately by multi-scale FCN decoders and the point cloud's Nearest Neighbor. To align the upscaled volumetric image features $\tilde{F}_l^{img}$ with the point cloud features $\hat{F}_l^{pt}$, we perform point-to-voxel mapping by sparsely sampling the image features at the point cloud coordinates $x^{pt}$. Next, two linear classifiers are applied to the cascaded point-wise image features $\{ \hat{F}_l^{img} \}_{l=1}^L$ and point cloud features $\{ \hat{F}_l^{pt} \}_{l=1}^L$ to classify points based on both image texture and edge structure, and produce the semantic predictions $P^{img}$ and $P^{pt}$.\\
\textbf{Cross-Modal Knowledge Distillation.}
In lower part of Fig. \ref{SMC-UDA_framework}, the multi-modal learning extracts edge structure within texture, so the point-wise image features $\hat{F}_l^{img}$ improve the point cloud structure representation by the fusion-then-distillation manner in each scale \cite{yan20222dpass}. The multi-modal fusion module fuses $\hat{F}_l^{img}$ and $\hat{F}_l^{pt}$, and conducts unidirectional alignment between image and edge. For point cloud enhancement, the $\hat{F}_l^{pt}$ is combined with $\hat{F}_l^{pt2img}$ through a skip connection to get the enhanced feature $\hat{F}_l^{ptE}$. After the MLP layer named "$\textsl{Fuse}$ Learner", we obtain the enhanced image feature:
\begin{equation}\hat{F}_l^{fuseE}=\hat{F}_l^{img}+\sigma (\textrm{MLP}(\hat{F}_l^{fuse})) \odot \hat{F}_l^{fuse}\end{equation}
Two independent classifiers, consisting of 2 linear layers and ReLU activation, are applied to the enhanced features $\hat{F}_l^{fuseE}$ and $\hat{F}_l^{ptE}$ to produce semantic segmentation scores $S_l^{fuse}$ and $S_l^{pt}$. The Kullback-Leibler ($\textrm{K-L}$) divergence as the unidirectional distillation loss drives structure score $S_l^{pt}$ to approach $S_l^{fuse}$, improving the generalization of the point cloud branch to deconstruct more different textures of images, which assess the correlation between the two branches:
\begin{equation}L_{xM}=D_{KL}\ (S_l^{fuse}\ ||S_l^{pt})= - \sum_{c\in C} s_l^{fuse}(c)\ \mathrm{log} \frac{s_l^{fuse}(c)}{s_l^{pt} (c)}
\end{equation}\\
\textbf{Supervised Learning on Source Domain. }
The proposed edge-based point cloud segmentation model involves supervised multi-modal training on source-domain data, which takes the image $x_s^{img}$ and point cloud coordinates $x_s^{pt}$ as inputs, with supervision from the point-wise label $y_s^{pt}$  as shown in the orange portion of Fig.\ref{SMC-UDA_framework}. The accuracy of point cloud segmentation $P_s^{pt}$ is maximized by a weighted cross-entropy and Lovasz-Softmax loss \cite{berman2018lovasz} defined as:
% Class probabilities $p_i(c)$ for each point $(i\in [0,N])$ construct the vector of pixel errors $m(c)$ for class $c\in [0,C]$,  $m(c)$ are
\begin{equation}
 L_{lvsz}= \frac{1}{C}\sum_{c\in C}{\overline{\bigtriangleup_{J_{c}}}}(\mathbf{m}_{c}), \ \mathrm{and} \ m_{i}(c)=\left\{\begin{array}{cc}
1-p_{i}(c), & if\  c=y_{i}(c)\\ p_{i}(c), & otherwise
\end{array}\right.
\end{equation}

where $p_i(c)$ is the class probability for each point $(i\in [0,N])$ with class $c\in [0,1]$ respectively. $\overline{\bigtriangleup_{J_{c}}}$ defines the Lovász extension of the Jaccard index, and $m_{i}(c)$ represents pixel errors on $y_i(c)$ which is the point-wise label of $y_s^{pt}$. We address class imbalance by using weighted cross-entropy, defined as $w_c=1/\sqrt{f_c}$, where $f_c$ represents class frequency: 
\begin{equation}
L_{wce}=-\frac{1}{(\sum_{c\in C}w_c )} \sum_{c\in C}{w_c y_c \textrm{log}(p_c)}
\end{equation}

The point cloud segmentation loss function is $L_{seg}=L_{lvsz}+L_{wce}$, with additional restricting intermediate predictions of multi-scale layers controlled by $\lambda_{s1}$ and $\lambda_{s2}$. The complete point cloud segmentation loss is formulated as: 
\begin{equation}\begin{array}{ll}
{L_{SEG}}_s(y_s^{pt})=& L_{seg}(P_s^{pt},y_s^{pt})+L_{seg}(P_s^{img},y_s^{pt})+ \\ 
&\sum_{l\in L}[\lambda_{s1} L_{seg}(S_l^{fuse},y_s^{pt})+\lambda_{s2}L_{seg}(S_l^{pt},y_s^{pt})]
\end{array}\end{equation}

The source domain loss is with a proportion $1: 0.05$ of $L_{seg}$ and $L_{xM}$, which is designed to stabilize the segmentation branch for self-constraint in multi-modal learning:
\begin{equation}
L_s=1\times{L_{SEG}}_s+0.05\times {L_{xM}}_s
\label{eq_ls}
\end{equation}
%the training on source domain with fully supervised annotation. 

\subsection{SMC-UDA}

The proposed self-training utilizes structure-constraint learned from the source domain, which self-supervises target domain in two ways. First, the structural predictions of point cloud $P_s^{pt}$ are obtained directly from the coordinates and used to constraint image structure learning while resisting texture noise \cite{yan2022let}. Secondly, the multi-scale predictions are supervised by $P_s^{pt}$ as the structural and shallow-to-deep supervision \cite{contrastivedeepsuper}. Source supervision is designed to learn rapidly and stably to guide the target domain learning, so the optimization is controlled by $\lambda_t = 0.01$ for a slow distillation. Specifically, the target domain's point cloud segmentation loss is formulated as follows:
\begin{equation}
    \begin{array}{ll}
        {L_{SEG}}_t\left(P_t^{pt}\right)=L_{seg}\left(P_t^{img},P_t^{pt}\right)+ \\
         \sum_{l\in L} \left[\lambda_1 L_{seg}\left(S_l^{fuse},P_t^{pt}\right)+\lambda_2 L_{seg} \left(S_l^{pt},P_t^{pt}\right)\right]
    \end{array}
\end{equation}

\begin{equation}
    L_t=\lambda_t \times (1\times {L_{SEG}}_t + 0.05 \times {L_{xM}}_t)
\label{eq_lt}
\end{equation}

\begin{figure}
\centering
\includegraphics[scale=.4]{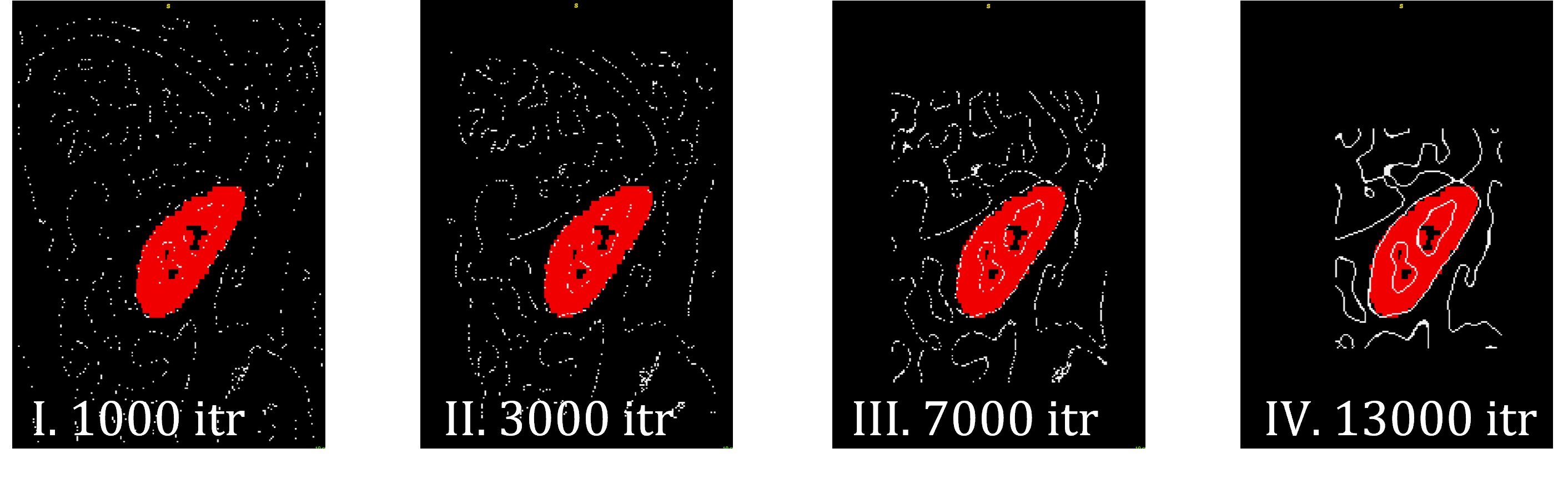}
\caption{A diagram of our progressive ROI sampling case. The sampled point cloud is too spare to contain enough structural information in a large ROI. With progressive ROI the sampler gradually focuses on the kidney area and increases the density of target organs, to provide more details of edge structure representation.}
\label{proroi}
\end{figure}

The SMC-UDA framework is trained in a domain-alternate end-to-end manner as Algorithm \ref{alg:SMC-UDA}. Each iteration includes supervised training on the source domain and unsupervised training on the target domain. To mitigate the sparse sampling, we propose progressive ROI as the demonstration in Fig. \ref{proroi}. The progressive ROI updates the sampled bounding boxes during validation epochs. Based on $y_{s}^{pt}$ and $P_t^{pt}$, the progressive iteration discards the edge map within 1/10 margins each time between the predicted target organ and the last sampling boundaries. Compared with the methods in \cite{zhong2019attention} with models having coarse and fine stages, our end-to-end model can learn from coarse to fine with different degrees of sparsity, making the training and testing more stable and robust.

\begin{algorithm}[!t]
    \caption{SMC-UDA algorithm}
    \label{alg:SMC-UDA}
    \renewcommand{\algorithmicrequire}{\textbf{Input:}}
    \renewcommand{\algorithmicensure}{\textbf{Output:}}
    \begin{algorithmic}[1]
        \REQUIRE Labeled source data $\mathit{D_{s}}$, unlabeled target data $\mathit{D_{t}}$. Maximum epoch $\textrm{M}$, validation epoch $\textrm{V}$.  %%input
        \ENSURE Final model parameters $\theta_{M}$, target domain predictions $\left\{ {P_t^{pt}}_{m^*} \right\}_{m^*=1}^{\left \lfloor M/V \right \rfloor}$     %%output
           
        \FOR{each $m \in [1,M]$}
            \STATE $\mathbf{Source\  domain\  learning}$:
            \begin{ALC@g}
                \STATE Sparse sample on $\mathit{D_{s}}$ within ROI
                \STATE Compute outputs and calculate $L_s$ based on Eq. \ref{eq_ls}
            \end{ALC@g}
            
            \STATE $\mathbf{Target\  domain\ adaptation}$:
            \begin{ALC@g}
                \STATE Sparse sample on $\mathit{D_{t}}$ within ROI
                \STATE Compute outputs and calculate $L_t$ based on Eq. \ref{eq_lt}
            \end{ALC@g}
            \STATE Calculate $L = L_s + L_t$
            \STATE Optimize and Update $\theta_{M}$
            \IF {$m = V$}
                \STATE Evaluate the target dataset
                \STATE Save target domain predictions ${P_t^{pt}}_{m^*}$
                \STATE Calculate boundary distance, update point cloud ROI
            \ENDIF
        \ENDFOR
        % \RETURN optimized $\theta_{M}$
    \end{algorithmic}
\end{algorithm}

\begin{table}[!t]
    \caption{\label{tab1} Performance comparison on kidney segmentation with Non-DA and UDA.}
    \centering
    \begin{tabular}{c|l|l|l|l|l|l|l|l|l}
    \toprule[ 1.3 pt]
        \multicolumn{10}{c}{KiTS21-CT → AMOS-CT} \\ 
        \cline{1-10}
        \multirow{2}{*}{Methods} & \multicolumn{3}{c|}{Dice(\%)↑} & \multicolumn{3}{c|}{ASSD(mm)↓} & \multicolumn{3}{c}{KD-Bbox-IoU(\%)↑} \\
        \cline{2-10}
          & L-KD & R-KD & Mean & L-KD & R-KD & Mean & L-KD & R-KD & Mean \\ \hline
        nnUNet-NoDA \cite{nnunet} & $\mathbf{78.1}$  & 77.5  & 77.8  & $\mathbf{3.6}$  & $\mathbf{2.8}$  & $\mathbf{3.2}$  & 76.1  & $\mathbf{80.8}$  & $\mathbf{78.5}$  \\ 
        PC branch (ours) & 76.0  & 77.6  & 76.8  & 4.6  & 4.5  & 4.6  & 74.2  & 75.1  & 74.7  \\ 
        Multi-modal (ours) & 77.7  & $\mathbf{79.1}$   & $\mathbf{78.4}$   & 4.4  & 4.6  & 4.5  & $\mathbf{76.9}$  & 78.1  & 77.5  \\ \hline
        PnP-AdaNet \cite{PnPAdaNet} & 45.5  & 45.3  & 45.4  & 10.1  & 9.6  & 9.9  & 46.4  & 49.0  & 47.7 \\ 
        SIFAv1 \cite{sifa1} & 42.0 & 48.0  & 45.0  & 17.5  & 16.6  & 17.0  & 13.8  & 26.0  & 19.9   \\ 
        SIFAv2 \cite{sifa2} & 80.8  & 84.0  & 82.4  & 3.7  & 2.8  & 3.2  & 65.4  & 75.1  & 70.2   \\ 
        % DAR-UNet \cite{darunet} & $\mathbf{87.2}$  & $\mathbf{89.5}$  & $\mathbf{88.4}$  & $\mathbf{2.4 }$ & $\mathbf{1.8 }$ & $\mathbf{2.1 }$ & 76.7  & $\mathbf{83.7}$  & 80.2  \\ 
        SMC-UDA (ours) & 81.2  & 82.3  & 81.8  & 3.2  & 3.3  & 3.2  & $\mathbf{80.3}$  & 81.2  & $\mathbf{80.8}$  \\ 
        \toprule[ 1.3pt]
        \toprule[ 1.3pt]
        \multicolumn{10}{c}{KiTS21-CT → CHAOS-MRI}  \\ \hline
        \multirow{2}{*}{Methods} & \multicolumn{3}{c|}{Dice(\%)↑} & \multicolumn{3}{c|}{ASSD(mm)↓} & \multicolumn{3}{c}{KD-Bbox-IoU(\%)↑} \\
        \cline{2-10}
         & L-KD & R-KD & Mean & L-KD & R-KD & Mean & L-KD & R-KD & Mean \\ \hline
        nnUNet-NoDA \cite{nnunet} & 0.0  & 0.3  & 0.2  & 110.2  & 81.7  & 89.8  & 0.0  & 0.3  & 0.1  \\ 
        PC branch (ours) & 78.8  & $\mathbf{90.3}$  & $\mathbf{84.5}$  & 5.8  & $\mathbf{1.7 }$ & $\mathbf{3.8 }$ & 72.1  & $\mathbf{84.6}$  & 78.4  \\ 
        Multi-modal (ours) & $\mathbf{82.3}$  & 86.3  & 84.3  & $\mathbf{4.5 }$ & 3.5  & 4.0  & $\mathbf{77.4}$  & 82.7  & $\mathbf{80.1}$  \\ \hline
        PnP-AdaNet \cite{PnPAdaNet} & 63.9  & 71.9  & 67.9  & 10.7  & 9.4  & 10.1  & 38.7  & 36.9  & 37.8 \\ 
        SIFAv1 \cite{sifa1} & 45.0  & 72.2  & 58.6  & 30.1  & 12.3  & 21.2  & 11.7  & 35.9  & 23.8 \\ 
        SIFAv2 \cite{sifa2} & 80.4  & 78.1  & 79.3  & 6.0  & 4.3  & 5.1  & 58.5  & 62.0  & 60.2   \\ 
        % DAR-UNet \cite{darunet} & 84.3  & $\mathbf{89.4}$  & 86.8  & 5.3  & 4.1  & 4.7  & 57.2  & 62.4  & 59.8  \\ 
        SMC-UDA (ours) & $\mathbf{89.0}$  & 87.3  & $\mathbf{88.1}$  & $\mathbf{1.9}$  & $\mathbf{1.6}$  & $\mathbf{1.8}$  & $\mathbf{84.1}$  & $\mathbf{87.2}$  & $\mathbf{85.7}$ \\
        \toprule[ 1.3pt]
    \end{tabular}
\end{table}
\section{Experiments}
\subsubsection{Dataset Settings.}
\begin{figure}
\centering
\includegraphics[scale=.33]{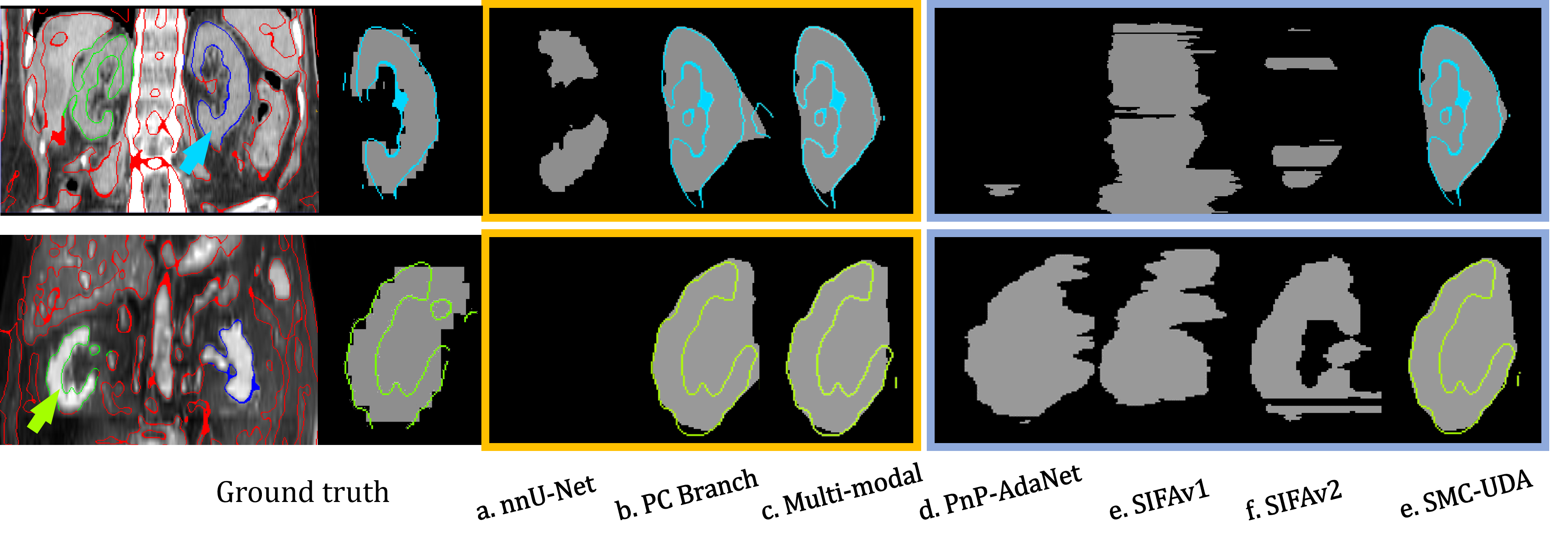}
\caption{Visualization of our proposed method and SOTA methods. The upper block contains the results of "KiTS21-CT → AMOS-CT" and the lower one contains "KiTS21-CT → CHAOS-MRI". The colored lines in b, c and h are the kidney predictions of edge maps. } 
\label{resulta}
\end{figure}

We evaluate the proposed SMC-UDA framework on cross-site and cross-modality kidney segmentation with three public datasets. Specifically, the source domain data is the largest open-source CT kidney dataset from the KiTS 2021 \cite{kits19} with 300 images and kidney annotations. For the target domain of the cross-site experiment, we use the CT data from a different institution, including 297 CT volumes from AMOS 2022 \cite{amos22}. For the target domain of the cross-modality experiment, we utilize the T2-SPIR MRI data with 20 volumes from CHAOS 2019 \cite{CHAOS19}. Each target domain dataset consists of abdominal multi-organ annotations, and we preserve the kidney annotation as the ground truth mask. To obtain a similar field of view for all volumes, the original scans were cropped at the abdominal volumes and scaled to $1\times1\times1$ voxel spacing. \\
\textbf{Model Implementation.}
As a pre-processing procedure, we normalize the intensity of the volumes and use the Difference of Gaussian \cite{edgedetectheory} as edge detection to obtain the edge maps. Due to memory limitations, we split volumes and process one side of the kidneys in every step, and randomly sample 100000 voxels on each edge map to obtain point clouds. The training batch sizes are 2 and 4 for source and target domains. Because of image size differences, the programs process batch data one by one with gradient accumulation. Our framework is implemented by PyTorch \cite{pytorch} using an RTX 3090 GPU and is optimized using the Adam optimizer. The maximum iteration is 15000 with an initial learning rate of $2 \times 10^{-4}$, which decayed with a ratio of 0.1 at 9000 and 12500 iterations. For every 1000 iterations, we validate the target domain and update the next progressive ROI, while collecting the point cloud predictions. We reconstruct the kidney mask from the accumulation of edge predictions with Open3D library \cite{open3d}. Alpha-shape algorithm \cite{alphashape} reconstructs the minimal watertight mesh of the predicted kidney boundary, and we fill the spatial hole of the voxelized meshes.  \\
\textbf{Model Comparison.} To evaluate the performance of our model, we compare it to the state-of-the-art UDA methods, including PnP-Adanet \cite{PnPAdaNet}, SIFAv1 \cite{sifa1}, SIFAv2 \cite{sifa2}, which are 2D generation-based adaptation methods. In a no-adaptation (Non-DA) setting, we compare the SOTA segmentation model nnUNet \cite{nnunet} with our multi-modal backbone and its point cloud branch (PC branch), which means these models are only trained with source domain data. The compared visualizations of kidney segmentation results are shown in Fig. \ref{resulta}. \\
The quantitative results are shown in Table \ref{tab1}. We use the Dice score and average surface distance (ASSD) as the segmentation metrics. The IoU scores of the kidney boundary box are unaffected by the different annotation of the renal pelvis, which illustrated in ground truth of Fig \ref{resulta}. Without domain adaptation, our edge-based model and point cloud branch can achieve similar performance with the SOTA methods, demonstrating the success of our edge structure learning. In the cross-modality experiment, the non-DA baseline can remarkably outperform nnUNet. In the UDA, our model outperforms the 2D-based UDA methods \cite{PnPAdaNet,sifa1,sifa2}. Without manual imaging alignment, the performance of generation-based UDA methods is highly limited by the quality of the generated image. The proposed SMC-UDA achieves an overall voxel segmentation Dice of 88.1\% and ASSD of 1.8 mm on the CHAOS-MRI data, highlighting the cross-domain effectiveness of the structure-constraint. With fewer false positives, our prediction demonstrates greater precision, establishing superiority over generative UDA models for kidney detection, as evidenced in IoU scores. 

\section{Conclusion}
In this paper, we investigate an edge structure constraint on cross-domain adaptation and propose a novel UDA segmentation method with multi-modal learning. To this end, we designed a two-branch multi-modal learning backbone, and trained with a cross-modal unidirectional distillation loss to the image texture and edge structure in the task of 3D kidney segmentation. The cross-domain self-training applied knowledge transfer between the predictions of the two domains and enforces the kidney structure constraint. With extensive experimental evaluations, our edge-based model achieves consistently improved performance of generalizability even without domain adaptation. Based on our UDA results, we are confident that the SMC-UDA can compete with the SOTA generation-based UDA methods by more effectively, learning cross-domain structure to realize further improvement of the generalizable performance.%  and can be applied to more challenging medical modalities

% limitations:
% One organ at a time
% Edge-based segmentation methods can be particularly useful for images that have a high degree of contrast and clear boundaries between structures.

% \subsubsection{Acknowledgements} We acknowledge the xxx (xxx-xxx-xxx xxxxxx, project xxxx),  and the xxx xxxx xxxx Research Centre.

\bibliographystyle{splncs04}
\bibliography{ref}

\end{document}